\documentclass[11pt]{article}
\pdfoutput=1

\usepackage[]{acl}

\usepackage{times}
\usepackage{latexsym}

\usepackage[T1]{fontenc}

\usepackage[utf8]{inputenc}

\usepackage{microtype}

\usepackage{enumitem} 
\usepackage{xurl}
\usepackage{graphicx}

\title{Ethics Sheets for AI Tasks}

\author{Saif M. Mohammad\\
	    National Research Council Canada\\
	     {\tt saif.mohammad@nrc-cnrc.gc.ca} }

\begin{document}
\maketitle
\begin{abstract}
Several high-profile events, such as 
the mass testing of emotion recognition systems on vulnerable sub-populations
and using question answering systems to make moral judgments,
have highlighted how technology will often lead to more adverse outcomes for those that are already marginalized.
At issue here are not just individual systems and datasets, but also the AI tasks themselves.
In this position paper, I make a case for thinking about ethical considerations not just at the level of individual models and datasets, but also at the level of AI tasks.
I will present a new form of such an effort, \textit{Ethics Sheets for AI Tasks}, 
dedicated to fleshing out the assumptions and ethical considerations hidden 
in how a task is commonly framed and 
in the choices we make regarding the data, method, and evaluation. 
I will also present a template for ethics sheets with 50 ethical considerations,
using the task of emotion recognition as a running example. 
Ethics sheets are a mechanism to engage with and document ethical considerations \textit{before} building datasets and systems.
Similar to survey articles, a small number of 
ethics sheets can serve numerous researchers and developers. 

\end{abstract}

\section{The Case: Importance of Ethics Considerations at the Level of AI Tasks}

 \setitemize[0]{leftmargin=5mm}
 \setenumerate[0]{leftmargin=8mm}

Good design helps everyone. It is well established, for example, that designing for accessibility helps society at large.\footnote{ https://blog.ai-media.tv/blog/why-designing-for-accessibility-helps-everyone} As Artificial Intelligence (AI), Machine Learning (ML), and Natural language Processing (NLP) systems become more ubiquitous, their broad societal impacts are receiving more scrutiny than ever before. However, several high-profile instances 
such as 
face-recognition systems that perform poorly for people with dark skin tones \cite{buolamwini2018gender},
machine translation systems that are biased against some genders \cite{prates2019assessing},
question answering systems that produce moral judgments \cite{talat2021word},
and 
mass testing of emotion recognition systems on certain sub-populations \cite{article19_2021,wakefield_2021},  
have highlighted how technology is often at odds with the very people it is meant to help, and how it will often lead to more adverse outcomes for those 
already marginalized. This raises uncomfortable questions for us AI researchers, developers, and leaders of technology companies:\\[-16pt]
\begingroup
\addtolength\leftmargini{-3mm}
\begin{quote}
{\it What role do we 
play in the 
harms perpetrated by technology?\\[2pt]
What are the 
assumptions in our research?
What are the implications of our choices?\\[2pt]
Are we striking at the barriers to opportunity or are we amplifying societal inequities?}
\end{quote}
\endgroup
\vspace*{-1mm}
\noindent The answers are often complex and multifaceted. 
While many AI systems have clear benefits, we are increasingly seeing
examples such as those discussed above where
real-world AI systems are causing harm.
Academic research (which often feeds into real-world systems), is
 also seeing growing amounts of criticisms:
criticisms  of physiognomy, racism, bias, discrimination, perpetuating stereotypes, 
ignoring indigenous world views,
and more.
See \citet{physiognomy_2017}
and \citet{ongweso_2020}
for recent examples.
There have also been 
criticisms of thoughtlessness 
(e.g., \textit{is automating this task, this way, really going to help people?}) and a seemingly callous disregard for the variability and complexity of human behavior \cite{mcquillan2018data,fletcher2018diversity,Birhane2021impossibility}.

This position paper makes the following contributions:
(1) It describes recent efforts by the AI community to 
encourage responsible research, the limitations of those efforts, and the need for
thinking about ethical considerations at the level of AI tasks. 
(2) Presents a detailed proposal for a new kind of document, \textit{Ethics Sheets for AI Tasks},
dedicated to fleshing out the assumptions and ethical considerations hidden 
in how a task is commonly framed and 
in the choices we make regarding the data, method, and evaluation.
(3) Provides a template for ethics sheets (that includes fifty ethical considerations),
with the task of automatic emotion recognition (AER) as a running example.  

NLP tasks, such as AER from text, machine translation, and summarization, are particularly rife with ethical considerations because they deal with language and people. Ethics sheets can help in recognizing and communicating the social and psychological complexities of language use; thereby, driving the desired design choices in NLP systems.   
More broadly, all AI tasks that deal with people and their artifacts (such as text, images, and video) can benefit from carefully thought out ethics sheets.
Every year, tens of thousands of people are joining the ranks of AI researchers and developers. 
Ethics sheets can serve them and others as useful introductory documents for AI tasks, 
guiding research/system design, facilitating the creation of datasheets and model cards, and
acting as springboards for new ideas in responsible research.

\subsection{Innovations for Responsible Research}
If a team builds a new dataset, then it is recommended to create a datasheet or data statement \cite{Gebru2018DatasheetsFD,bender-friedman-2018-data}
that lists key details of the dataset such as composition and intended uses. It is meant to encourage appropriate use of the data.
If a team builds a new system, then it is recommended to create a model card \cite{mitchell2019model}
that lists key details of the model such as performance in various contexts and intended use scenarios. It is meant to encourage appropriate use of the system.
For individual papers, we write ethics/impact statements; and conferences have started to institute ethics policies and ethics reviews.\\[3pt] 
\noindent {\bf Limitations:} Datasheets and model cards are pivotal inventions that will serve our community well. However, they are not without limitations and the specificity of their scope (on individual pieces of work) places additional constraints:\\[-16pt]
\begin{itemize}[itemsep=-3pt]
\item Authors are in a position of conflict of interest; there are strong incentives to present their work in positive light (for paper acceptance, community buy-in, etc.)
\item There can be a tendency to produce boiler-plate text without a meaningful and critical engagement with the relevant ethical issues.
\item While there is important benefit in creating post-production documents that describe societal impact, it is arguably more important to engage with ethical considerations (and publish an ethics focused document) \textit{before} building AI systems (and possibly even choosing to not build a system for a particular deployment context based on the analysis).
\item Lastly, ethics considerations apply at levels other than individual projects; e.g., 
at the level of AI tasks. A comprehensive engagement with the relevant ethical issues requires a wide literature review, and the resulting analysis to be presented in a dedicated document (and not in add-on sections for individual system papers).
\end{itemize}

\subsection{Ethics at the Level of AI Tasks}
 I am defining \textit{AI task} to simply mean some task we may want to automate using AI techniques. An \textit{AI system} is a particular AI model built for the task. Individual systems have their own unique sets of ethical considerations (depending on the choices that were made when building the systems). However, several ethical considerations apply not at the level of individual systems, but at the level of the task. For example, consider the task of detecting personality traits from one's 
 utterances. Even before we consider 
 a system for the task, we ought to consider questions such as:\\[-17pt]
\begin{itemize}[itemsep=-3pt]
\item What are the societal implications of automating personality trait detection?
\item How can such a system be used/misused?
\item Is there enough credible scientific basis for personality trait identification that we should attempt to do this?
\item Which theory of personality traits should such automation rely on? What are the implications of that choice?
\end{itemize}
\vspace*{-1mm}
\noindent And so on. 
In addition, 
for a given task, there exist ethical considerations latent in the choices commonly made in  dataset creation, model development, and evaluation. Poor choices 
lead to more harm. Consider these outcomes reported in the popular press:\\[-17pt]
\begin{itemize}[itemsep=-3pt]
\item Text Generation: \textit{`Dangerous' AI writes fake news}, BBC.\footnote{www.bbc.com/news/technology-49446729} 
\item Image Generation: \textit{`Deepfakes' a political problem already hitting  EU}, EU Observer.\footnote{https://euobserver.com/opinion/151935}\\[-12pt]
\item Automatic Emotion Recognition from Faces: 
\textit{ 
China's emotion recognition market and its implications for human rights}, Article19 \footnote{www.article19.org/wp-content/uploads/2021/01/ER-Tech-China-Report.pdf}.\\[-12pt]
\item Machine Translation: 
\textit{Female historians and male nurses do not exist, Google Translate tells its European users}, Algorithm Watch.\footnote{\resizebox{0.44\textwidth}{!}{https://algorithmwatch.org/en/google-translate-gender-bias}}\\[-12pt]
\item Information Extraction: \textit{Google apologises for `ugliest Indian language' search result}, BBC.\footnote{www.bbc.com/news/world-asia-india-57355011}\\[-12pt]
\end{itemize}
\vspace*{-2mm}
\noindent Numerous other such examples have surfaced in just the past few years for a variety of AI tasks. 

Additionally, fields such as NLP and Computer Vision organize themselves in sub-fields by task (e.g., machine translation). 
Laws about AI ethics are also emerging in the context of AI tasks \cite{european2020artificial} -- e.g., based on whether the task is high risk.
Reading relevant literature, engaging with stakeholders, and past experience in developing systems helps one to start identifying relevant ethical considerations for 
an AI task;
but that takes time.  Meanwhile, 
tens of thousands of new researchers are joining our ranks. 
Pressures to graduate and find good jobs force them to build systems and publish papers in a matter of months.
Even experienced researchers can find it difficult to keep track of various ethical considerations discussed in a wide assortment of conferences and journals. 


\section{Proposal: Ethics Sheets for AI Tasks}

If one wants to do work on an AI Task, then right at the beginning it is useful to have access to:\\[-17pt]
\begingroup
\addtolength\leftmargini{-3mm}
\begin{quote}
    \textit{a document that substantively engages with the ethical issues relevant to that task; going beyond individual systems and datasets, drawing on a body of relevant work.}
\end{quote}
\endgroup
\vspace*{-1mm}
\noindent Similarly, if one conceptualizes a new AI Task, then 
it is useful to simultaneously create such
a source of information. 

Therefore, I propose that we researchers and developers write such articles, which I will refer to as \textit{Ethics Sheets for AI Tasks}.
In some ways, ethics sheets are similar to survey articles for areas of research, except here the focus is on ethical considerations for an AI task.
Simply put: an ethics sheet for an AI task is a semi-standardized article that aggregates and organizes a wide variety of ethical considerations relevant for that task.
It:\\[-17pt]
\begin{itemize}[itemsep=-3pt]
    \item Fleshes out assumptions hidden in how the task is framed, and in the choices often made regarding the data, method, and evaluation.
    \item Presents ethical considerations unique or especially relevant to the task.
    \item Presents how common ethical considerations manifest in the task.
    \item Presents relevant dimensions and choice points; along with tradeoffs for various stakeholders.
    \item Lists common harm mitigation strategies.
    \item Communicates societal implications 
    to researchers, developers, and the broader public.
\end{itemize}
\noindent The sheet should flesh out various ethical considerations that apply at the level of task. It should also flesh out ethical consideration of common theories, methodologies, resources,  and practices used in building AI systems for the task.

Ethics sheets may sometimes suggest that certain applications in specific contexts are appropriate or inappropriate, but largely they are meant to 
discuss the various considerations to be taken into account when the developer is deciding 
whether to build or use a particular system, 
how to build it, and
how to assess its societal impact.
It is meant to help the developer identify what is more appropriate for their given deployment context. 

A good ethics sheet will question some of the assumptions that often go unsaid.
It will encourage more thoughtfulness:\\[-17pt] 
\begin{itemize}[itemsep=-3pt]
    \item {\it Why should we automate this task?
    \item What is the degree to which human behavior relevant to this task is inherently ambiguous and unpredictable?\\[-12pt]
    \item What are the theoretical foundations? 
    \item What social and cultural forces  motivate choices in task design, data, methodology, and evaluation?
(Science is not immune to these forces---there is no `view from nowhere').
    \item How is the automation of the task going to impact various groups of people?
    \item How can the automated systems be abused?
    \item Is this technology helping everyone or only those with power and advantage?} etc.\\[-16pt]
\end{itemize}
\noindent Thinking about these questions is important if we want to break away from the current paradigm of building things that are divisive (that work well for some and poorly for others) and instead move towards building systems that treat human diversity and variability as a feature (not a bug); systems that truly dismantle barriers to opportunity, and bring diverse groups of people together. Thus, questions such as those shown above can be useful in determining what is included in ethics sheets.\\[5pt]
\noindent {\bf Target audience:} The target audience 
of a sheet 
includes the various stakeholders of the AI Task. The stakeholders may or may not have the time and background to understand the technical intricacies of an AI task. However, they build on, use, and make laws about what we create.
Further, people are impacted by AI systems. They should be able to
understand its decisions that impact them,
understand its broad patterns of behaviour,
contest the predictions, and 
find recourse. Ethics sheets can help to that end.
It is our responsibility to describe our creations in accessible terms, so that others can make informed decisions about them.
Thus the target audience includes:\\[-16pt]
\begin{itemize}[itemsep=-3pt]
\item Researchers;  developers 
\item Educators (esp.\@ those who teach AI, ethics)
\item Policy makers; politicians
\item People whose data is used; society at large
\end {itemize}
\vspace*{-1mm}
\noindent Owing to differences in backgrounds and needs, it is better to create versions of the Ethics Sheet tailored to stakeholders, for example:\\[-16pt]
\begin{itemize}[itemsep=-3pt]
    \item One sheet for society at large (with a focus on how system behaviour can impact them and how they can contribute/push-back);
    \item One sheet for researchers, developers, and the motivated non-technical reader (with a greater emphasis on system building choices). 
\end{itemize}
\vspace*{-1mm}
Ethics sheets complement datasheets and model cards: 
while the latter are post-production documents produced by system/data builders, ethics sheets are meant to be accessed
before building systems. Similar to traditional survey articles, a small number of carefully created ethics sheets can serve numerous researchers and developers creating systems and data for AI tasks.

See the FAQ in the \hyperref[sec:appendix]{Appendix} (after references) for a discussion on some practicalities 
involved with who should create ethics sheets, when they should be created, for which tasks, etc. I discuss below
some key characteristics and benefits of ethics sheets, followed by 
a template and a pointer to an example ethics sheet in the next section.

\subsection{No One Sheet to Rule them All}

A single ethics sheet does not speak for the whole community (just as survey articles do not speak for the whole community).
No one 
group
 can claim authority or provide the authoritative ethics sheet for a task. Ethics sheets can be created through large community efforts (through workshops or carefully maintained wikis) and smaller individual and group efforts. 
Efforts led by small teams may miss important perspectives. However, community efforts face several logistical and management challenges. They also have the tendency to only include agreed upon non-controversial ideas that do not threaten existing power structures. While each of these approaches to implement ethics sheets has their pros and cons, a multiplicity of ethics sheets is likely most promising.
Multiple ethics sheets created (by different teams and approaches) 
reflect multiple perspectives, viewpoints, and what is considered important to different groups of people.
We should be wary of a world where we have single authoritative ethics sheets per task and no dissenting voices. 

\subsection{Work on Ethics a Perpetual Task} The set of ethical considerations for a task is not a static list; it needs to be continuously or periodically revisited and updated.
The considerations can be developed iteratively and organically, in small teams and in large community efforts (say through dedicated workshops). 
The ethics sheet is not a silver bullet to make things perfect, lead to easy solutions, or ``solve ethics''.
The goal is to raise awareness of relevant ethical considerations, encourage following of established best practices, and inspire new ideas of responsible research appropriate for one's particular context. 

\subsection{Components of an Ethics Sheet}

The sections below are central. However, individual tasks may warrant additional sections.\\[3pt]
\noindent {\bf Preface:} Present why and how the sheet came to be written. The process followed. Who worked on it along with their professional or lived experience relevant to the subject matter. Challenges faced in writing the sheet. Changes made, if a revision of an earlier sheet. Version number, date published, and contact information.\\[3pt]
{\bf Introduce, Define, Set Scope:} Introduce the task and some common manifestations of the task. Define relevant terminology. Set the scope of the ethics sheet (e.g., maybe you are creating a sheet for speech input, but not textual input).\\[3pt]
{\bf Motivations and Benefits:} Provide an overview of common benefits and motivations of the task.\\[3pt]
{\bf Ethical Considerations:} This is the star of the show. Aggregate and organize the ethical considerations associated with the AI task. Present the trade-offs associated with choices. Present harm mitigation strategies. Cite relevant literature. Organization of ethical considerations should be based on the primary target audience. For example, ethics sheets primarily for researchers and developers may benefit 
from sub-sections on:
Task Design, Data, Method, and Evaluation. Task design may benefit from sections for theoretical foundations and `why automate this task?'. Evaluation will benefit from sub-sections that go beyond quantitative metrics.\\[3pt]
{\bf Other:} Include anything that helps with the goals of the Ethics Sheet.

\subsection{Benefits of Ethics Sheets}
Ethics sheets for AI Tasks address a number of concerns raised in the first section of this paper.
Specifically, their benefits include:\\[-16pt] 
\begin{enumerate}[itemsep=-3pt]
     \item Encourages  thoughtfulness regarding why to automate, how to automate, and how to judge success 
     well before the building systems.
    \item Fleshes out assumptions in how the task is commonly framed, and in the choices often made regarding data, method, and evaluation.
    \item Presents the trade-offs of relevant choices so that stakeholders can make informed decisions appropriate for their context. 
    Ethical considerations often involve a cost-benefit analysis; where we draw the lines may differ depending on our cultural and societal norms. 
    \item Identifies points of agreement and disagreement. Includes multiple points of view.
    \item Moves us towards consensus and  standards.
    \item Helps us navigate system development choices.
    \item Helps develop better datasheets, model cards.
    \item Has citations and pointers; acts as a jumping off point for further reading.
    \item Helps stakeholders challenge assumptions made by researchers and developers.
    \item Helps stakeholders develop harm mitigation strategies.
    \item Standardized sections and a familiar look and feel make it easy for the compilation and communication of ethical considerations.
    \item Helps engage the various stakeholders of an AI task with each other.
    \item Multiple ethics sheets created for the same task reflect multiple perspectives, viewpoints, and what is considered important to different groups of people at different times.
    \item Acts as a great introductory document for an AI Task (complements survey articles and task-description papers for shared tasks). 
\end{enumerate}

\section{A Template and an Example Sheet}

I present below a template that can serve as a handy starting point in the creation of new ethics sheets,
and that further clarifies what can be included in an ethics sheet. 
In the template below I will use Automatic Emotion Recognition (AER) as the running example.
AER is a particularly interesting, widely applicable, and complex example of AI tasks with notable benefits and risks.
Thus an ethics sheet for AER can be particularly instructive.

In her seminal book, Affective Computing, Dr.\@ Rosalind Picard described Automatic Emotion Recognition (AER) as: ``giving emotional abilities to computers”. It is a sweeping interdisciplinary area of study exploring many foundational research questions and many applications \cite{picard2000affective}. However, some of the recent commercial and governmental uses of AER have garnered considerable criticism, including:
infringing on one's privacy, exploiting vulnerable sub-populations, and even allegations of downright pseudo-science \cite{wakefield_2021,article19_2021,woensel_nevil_2019}.
Even putting aside high-profile controversies, emotion recognition impacts people and thus entails ethical considerations (big and small). 
\citet{Mohammad21-Ethics-AER} presents an ethics sheet for automatic emotion recognition and sentiment analysis. It is a critical reflection of this broad field of study with the aim of facilitating more responsible emotion research and appropriate use of the technology.
 I will use some details from that sheet below to clarify the elements of the generic template.

\subsection{Preface}

The preface is an opportunity to frame the discussion. \citet{Mohammad21-Ethics-AER} presents rapid-fire questions such as whether it is ethical to do automatic emotion recognition, how automatic recognition can mean many things, and it can be deployed in many contexts, how emotions are particularly personal, private, and complex; and how the ethics sheet can help in more responsible AER research as well as responsible system development and deployment. 
It also lists the primary motivation for the ethics sheet and the target audience. 

\subsection{Modalities and Scope}

\noindent \textbf{Modalities:} AI tasks may involve various modalities. For example, work on AER has made use of
facial expressions, gait, 
skin conductance, blood conductance, 
force of touch, 
speech, written text, etc.
All of these modalities come with benefits, potential harms, and ethical considerations.\\[3pt]
\noindent \textbf{Scope:} Specifying the scope of an ethics sheet allows sharper focus. \citet{Mohammad21-Ethics-AER} focuses on AER from written text.

 \subsection{Task}

Clarify the task. \citet{Mohammad21-Ethics-AER} states that emotion recognition 
is a broad umbrella term used to refer to a number of related tasks such as inferring emotions the speaker is trying to convey, 
inferring patterns of speaker's emotions over longer periods of time, 
tracking impact of health interventions on one's well-being,
inferring speaker's attitudes/sentiment towards a target product, movie, person, idea, policy, entity, etc. Each of these framings has ethical considerations and may be more or less appropriate for a given context. For example, framing the task as determining the mental state 
is especially problematic due to concerns about privacy and reliability.

\subsection{Applications}

Discussing applications of the task is important 
not only because it is an opportunity to present the benefits of the task but also
because an understanding of the applications is crucial to recognizing various ethical considerations. 
\citet{Mohammad21-Ethics-AER} presents a sample of existing applications of AER in
public health, commerce, government policy, art and literature, research (social Sciences, neuroscience, psychology), and intelligence.
Note also that all of the benefits come with potential harms and ethical considerations. Use of AER for military intelligence and education is especially controversial and laced with ethical considerations. 

\subsection{Ethical Considerations}

The usual approach to building a system for an AI task is to design the task (e.g., for AER, identify the precise emotion task to be automated, identify the emotions of interest,  etc.), compile appropriate data (e.g., label some of the data), train ML model (method) to capture relevant patterns of language from the data, and evaluate the model by examining their predictions on a held-out test set. There are ethical considerations associated with each step of this development process.
Below is a template 
of 50 considerations grouped by the associated stage: Task Design, Data, Method, Impact, Privacy \& Social Groups (this final category is particularly important and cuts across Task Design, Data, Method, and Impact). 
I present only a high-level summary for each category below. 
See \citet{Mohammad21-Ethics-AER} for an instantiation of this generic template for the task of automatic emotion recognition (AER).
It includes details on how these considerations manifest in AER.
One can use the template below as a guide (in part or full), skip the considerations that do not apply, and 
describe how the relevant considerations manifest for their chosen task. One should notably include details of key considerations for their task
whether it is included in this template or not.
One can also cite specific issues already discussed in the ethics sheets for other tasks.\\[10pt]
\noindent \textbf{TASK DESIGN}\\[4pt]
\noindent \textit{Summary:} This section discusses various ethical considerations associated with the choices involved in the framing of the focus task and the implications of automating the focus task. For AER, important considerations included: whether it is even possible to determine one's internal mental state; whether it is ethical to determine such a private state; and who is often left out in the design of existing AER systems. \citet{Mohammad21-Ethics-AER} also discusses how it is important to consider which formulation of emotions is appropriate for a specific task/project; while avoiding careless endorsement of theories that suggest a mapping of external appearances to inner mental states.\\[4pt]
\textbf{A. Theoretical Foundations}\\[4pt]
1. \textit{Task Design and Framing:} Discuss notable task formulations and their ethical implications.\\[1pt]
2. \textit{Theoretical Models and their Implications:} Discuss notable theoretical constructs from linguistics, psychology, etc.\@ that underpin the focus AI task. Discuss the ethical considerations associated with these constructs.  \\[1pt]
3. \textit{Meaning and Extra-Linguistic Information:} Discuss how nuances of meaning in text, images, etc.\@ and extra-linguistic information play a role in the task;
and that systems that make use of limited information may lead to false predictions.\\[1pt]
4. \textit{Wellness and Health Implications:} Discuss implications of the task design on wellness and health of people (if any). \\[1pt]
5. \textit{Aggregate Level vs. Individual Level Prediction:} Discuss whether the goal is to determine something about individuals or groups of people, how that choice impacts the ethical considerations associated with the task.\\[4pt]
\noindent \textbf{B. Implications of Automation}\\[4pt]
\noindent  6. \textit{Why Automate:} Discuss who benefits from this automation; and whether 
this will shift power to those that need it the most \cite{kalluri2020don}.\\[1pt] 
7. \textit{Embracing Diversity:} Discuss how design choices impact diverse groups of people. \\[1pt]
8. \textit{Participatory/Emancipatory Design:} Discuss how people that are impacted by the technology can play a role in shaping task design.\\[1pt]
9. \textit{Applications, Dual Use, Misuse:} Discuss how task design can enhance applications. 
Discuss prohibited and contentious use case scenarios.
Discuss how task design can mitigate some of the harms associated with the task. (Note that even when systems are used as designed, they can lead to harm.)\\[1pt]
10. \textit{Disclosure of Automation:} Discuss the ethical ramifications of disclosing and of not disclosing to the users that the underlying task is automated.\\[10pt]
\noindent \textbf{DATA}\\[4pt]
\noindent \textit{Summary:} This section has three broad themes: implications of using datasets of different kinds, the tension between human variability and machine normativeness, and the ethical considerations regarding the people who have produced the data. Notably, \citet{Mohammad21-Ethics-AER} discusses how on the one hand is the tremendous variability in human 
representation and expression of language and emotions, and on the other hand, is the inherent bias of modern machine learning approaches to ignore variability. Thus, through their behaviour (e.g., by recognizing some forms of emotion/language expression and not recognizing others), AI systems convey to the user what is ``normal"; implicitly invalidating other forms of emotion/language expression.\\[4pt]
\textbf{C. Why This Data}\\[4pt]
11. \textit{Types of data:} Discuss notable types of data such as labeled training data, large internet-scraped raw data for language models, lexicons, image repositories, etc.\@ and their ethical implications.\\[1pt]
12. \textit{Dimensions of data:} Discuss notable dimension of data such as size, whether it is carefully curated for the research or uncurated data obtained from an online platform, less private/sensitive data or more private/sensitive data, what languages are represented in the data, degree of documentation provided with the data, and so on. \\[4pt]
\noindent \textbf{D. Human Variability--Machine Normativeness}\\[4pt]
13. \textit{Variability of Expression, Conceptualization:} Discuss how variability of human expression (e.g., in text, images, videos, etc.) and representations of meaning impacts the associated task. \\[1pt]
14. \textit{Norms of Emotions Expression:} Discuss how some task-associated forms of human expression may be considered "normal" or "correct" by a group of people, and the extent to which other forms of expression are also valid and appropriate. Discuss how systems for the task are impacted by various design, data, and method choices when it comes to recognizing various forms of appropriate expressions.\\[1pt] 
15. \textit{Norms of Attitudes:} Discuss how different people may have different attitudes towards other people and entities (some of which may be inappropriate), and how AI systems for the task may produce responses laden with such attitudes.\\[1pt]
16. \textit{"Right" Label or Many Appropriate Ones:} Discuss whether for the given task, certain training instances can/should be labeled with multiple appropriate responses. Discuss implications of choices such as keeping only the majority label from the annotators.\\[1pt]
17. \textit{Label Aggregation:} Discuss notable approaches to label aggregation, and their implications. (See \citet{aroyo2015truth,checco2017let}.)\\[1pt]
18. \textit{Training on Historical Data:} Discuss implications of training systems on historical data; who is missing from the data; biases in the data.\\[1pt]
19. \textit{Training--Deployment Differences:} Discuss implications of deploying systems on data that is markedly different from the training data.\\[4pt]
\noindent \textbf{E. The People Behind The Data}\\[4pt]
20. \textit{Platform Terms of Service:} Discuss implications of relevant terms of services associated with platforms from which data was obtained.\\[1pt]
21. \textit{Anonymization, Ability to Delete One's Data:} Discuss importance of anonymization, and the ability to control/delete one's data.\\[1pt]
22. \textit{Warnings and Recourse:} Discuss appropriate levels of warnings and recourse one should provide when building and deploying systems.\\[1pt]
23. \textit{Crowdsourcing, Expert Annotation:} Discuss the implications of training AI systems on crowdsourced data and expert annotations. \\[10pt]
\noindent \textbf{METHOD}\\[4pt]
\noindent \textit{Summary:} Discuss the ethical implications of deploying a given method for the focus task. Present the types of methods and their tradeoffs, as well as considerations of who is left out and spurious correlations. 
\citet{Mohammad21-Ethics-AER} also discusses green AI and the fine line between emotion management and manipulation.\\[4pt]
\noindent \textbf{F. Why This Method}\\[4pt]
24. \textit{Types of Methods and their Tradeoffs}: Discuss how different methods entail different trade-offs, e.g.,
less accurate vs.\@ more accurate, white box 
vs.\@ black box, 
less data hungry vs.\@ more data hungry,
less privacy preserving vs.\@ more privacy preserving,
fewer inappropriate biases vs.\@ more inappropriate biases, etc.\\[1pt]
25. \textit{Who is Left Out by this Method:} Discuss whose voices tend to not be included because of the method and data used.\\[1pt]
26. \textit{Spurious Correlations:} Discuss the tendency and implications of the chosen method to rely on spurious correlations in the data. (See \citet{agrawal2016analyzing,bissoto2020debiasing}.)\\[1pt] 
27. \textit{Context is Everything:} Discuss how greater context can impact system accuracy and also the corresponding implications on privacy.\\[1pt]
28. \textit{Individual Expression Dynamics:} Discuss how variability and other characteristics of an individual's expression over time (e.g., their speech patterns) impact the task.\\[1pt]
29. \textit{Historical Behavior vs.\@ Future Behavior:} Discuss the extent to which past behavior is not indicative of future behavior, and the impact of methods that assume the contrary.\\[1pt]
30. \textit{Communication Management, Manipulation:} In case of human interaction systems, discuss whether the system is simply managing communication or if it can be used to nudge a person to a certain behavior.\\[1pt]
31. \textit{Green AI:} Discuss the energy implications of the chosen method \cite{Strubell_Ganesh_McCallum_2020,schwartz2020green}.\\[10pt]
\noindent \textbf{IMPACT AND EVALUATION}\\[4pt]
\noindent \textit{Summary:} This section discusses ethical considerations associated with the evaluation of the focus task systems (Metrics) as well as the importance of examining systems through a number of other criteria (Beyond Metrics). Notably, 
\citet{Mohammad21-Ethics-AER} discusses interpretability  
and contestability, because even when systems work as designed, there will be some negative consequences. Recognizing and planning for such outcomes is part of responsible development.\\[4pt]
\noindent \textbf{G. Metrics}\\[4pt]
32. \textit{Reliability/Accuracy:} Discuss commonly used (traditional) metrics for evaluating systems such as accuracy, F-score, and reliability. Discuss their limitations.\\[1pt]
33. \textit{Demographic Biases:} Discuss when and how systems can be unreliable or systematically inaccurate for certain groups of people, races, genders, people with health conditions, people from different countries, etc. (See \citet{buolamwini2018gender,kiritchenko-mohammad-2018-examining}.)\\[1pt]
34. \textit{Sensitive Applications:} Discuss whether systems for the task should be used in sensitive scenarios such as impacting one's health, livelihood, or freedom, and if such use is acceptable then under what conditions. Unless a clear case can be made for such uses, it is best to caution against such use of AI systems.\\[1pt]
35. \textit{Testing:} Discuss how systems should be tested on a diverse set of datasets and metrics.\\[4pt]
\noindent \textbf{H. Beyond Metrics}\\[4pt]
36. \textit{Interpretability, Explainability:} Discuss task-specific approaches to system interpretability and explainability of systems and their role in identifying biases and flaws.\\[1pt]
37. \textit{Visualization:} Discuss how suitable visualizations (especially interactive ones) can allow users to explore trends in the data and system behavior; and importantly, allow one to drill down to the source data that is driving the trends.  \\[1pt]
38. \textit{Safeguards and Guard Rails:} Discuss notable task-specific safeguards to prevent harm to individuals.\\[1pt]
39. \textit{Harms when the System Works as Designed:} Discuss how systems that work as designed can still cause harms.\\[1pt]
40. \textit{Contestability and Recourse:} Discuss best practises in allowing users to contest system predictions, and in terms of appropriate recourse.\\[1pt]
41. \textit{Ethics Washing:} Discuss how ethics documentation should be used to meaningfully engage with the issues rather than for cosmetic purposes.\\ [10pt]
\noindent \textbf{PRIVACY AND SOCIAL GROUPS}\\[4pt]
\noindent \textit{Summary:} 
The privacy section discusses both individual and group privacy. \citet{Mohammad21-Ethics-AER} points out how the idea of group privacy becomes especially important in the context of soft-biometrics determined through AER that are not intended to be able to identify individuals, but rather identify groups of people with similar characteristics. The subsection on social groups discusses the need for work that does not treat people as a homogeneous group (ignoring group differences and implicitly favoring the majority group) but rather values disaggregation and explores intersectionality, while minimizing reification and essentialization of social constructs.\\[4pt]
\noindent \textbf{I. Implications for Privacy}\\[4pt]
42. \textit{Privacy and Personal Control:} Discuss privacy implications of the task, and measures to give more control to the user on their data.\\[1pt]
43. \textit{Group Privacy and Soft Biometrics:} Discuss implications of automating the task on group privacy \cite{floridi2014open}.\\[1pt]
44. \textit{Mass Surveillance vs.\@ Right to Privacy, Freedom of Expression, Right to Protest:} Discuss implications of automating the task
on the ability to monitor behavior of a large number of people, and trade-offs with the right to privacy, freedom of expression, and the right to protest.\\[1pt]
45. \textit{Right Against Self-Incrimination:} Automating certain tasks may make it easy for systems to find incriminating information produced by an individual. This can work against the right afforded by many countries against self-incrimination. Discuss any pertinent considerations.\\[1pt]
46. \textit{Right to Non-Discrimination:} Discuss whether automating the task can be used to discriminate against certain groups of people. Discuss safe guards.\\[4pt]
\noindent \textbf{J. Implications for Social Groups}\\[4pt]
47. \textit{Disaggregation:} When building automatic prediction systems: Report performance disaggregated for each of the relevant and key demographic groups. (See work on model cards \citet{mitchell2019model}.) Cite work reporting disaggregated results for the task.)\\[1pt]
48. \textit{Intersectionality:} People with multiple group identities are often not seen as prototypical members of any of their groups and thus are subject to, what is refered to as, intersectional invisibility---omissions of their experiences in historical narratives and cultural representation, lack of support from advocacy groups, and mismatch with existing anti-discrimination frameworks. Discuss implications of the task on those with multiple group identities.\\[1pt]
49. \textit{Reification and Essentialization:}  Avoid reinforcing false beliefs that there are innate differences across different groups or that some features are central for one to belong to a social category. Appropriately contextualize work on disaggregation; for example, by impressing on the reader that even though constructs such as race are artificial and social in nature, the impact of people’s perceptions and behavior around race lead to very real-world consequences. \\[1pt]
50. \textit{Attributing People to Social Groups:} In order to be able to obtain disaggregated results, sometimes one  needs access to demographic information. 
This leads to considerations such as: whether the participants are providing meaningful consent to the collection of such data and whether the data is being collected in a manner that respects their privacy, their autonomy (e.g., can they choose to delete their information later), and dignity (e.g., allowing self-descriptions).


\section{Concluding Thoughts}

In this position paper, I discussed how ethical considerations apply not just at the level of individual models and datasets, but also at the level of AI Tasks.
I presented a new form of documenting ethical considerations, 
which I call \textit{Ethics Sheets for AI Tasks}. It is a document  
dedicated to fleshing out the assumptions and ethical considerations hidden 
in how a task is commonly framed and 
in the choices we make regarding the data, method, and evaluation.
I listed various benefits of such ethics sheets
and discussed caveats such as how a single ethics sheet does not speak for the whole community.
I also provided a template sheet and an example, proof-of-concept, ethics sheet for automatic emotion recognition.
Ethics sheets have the potential for engaging  various stakeholders of AI tasks towards responsible research and development.
I hope that this work spurs the wider community to ask and document:\\[6pt] 
 \hspace*{2mm} \textit{What ethical considerations apply to my task?} \\[-3pt] 

\noindent \textit{Note:} See FAQ in the \hyperref[sec:appendix]{Appendix} for practical
considerations involved in
who should create ethics sheets, when, for what tasks, etc.

\section*{Acknowledgments}

I am grateful to Annika Schoene, Isar Nejadgholi, Mohamed Abdalla, and Tara Small for encouragement on the initial idea of Ethics Sheets for AI Tasks, 
 the thoughtful discussions, and comments on earlier drafts. 
  Many thanks to Emily Bender, Esma Balkir, Patricia Thaine, Brendan O'Connor, Cyril Goutte, and Sowmya Vajjala for thoughtful comments on an early draft.
 Many thanks to Mallory Feldman, Roman Klinger, Rada Mihalcea, and Peter Turney for thoughtful comments on the ethics sheet for emotion recognition. 


\bibliography{acl2020}

\begin{thebibliography}{26}
\expandafter\ifx\csname natexlab\endcsname\relax\def\natexlab#1{#1}\fi

\bibitem[{Agrawal et~al.(2016)Agrawal, Batra, and
  Parikh}]{agrawal2016analyzing}
Aishwarya Agrawal, Dhruv Batra, and Devi Parikh. 2016.
\newblock Analyzing the behavior of visual question answering models.
\newblock \emph{arXiv preprint arXiv:1606.07356}.

\bibitem[{Arcas et~al.(2017)Arcas, Mitchell, and Todorov}]{physiognomy_2017}
Blaise Arcas, Margaret Mitchell, and Alexander Todorov. 2017.
\newblock \href
  {https://medium.com/@blaisea/physiognomys-new-clothes-f2d4b59fdd6a}
  {Physiognomy’s new clothes}.

\bibitem[{Aroyo and Welty(2015)}]{aroyo2015truth}
Lora Aroyo and Chris Welty. 2015.
\newblock Truth is a lie: Crowd truth and the seven myths of human annotation.
\newblock \emph{AI Magazine}, 36(1):15--24.

\bibitem[{ARTICLE19(2021)}]{article19_2021}
ARTICLE19. 2021.
\newblock \href
  {https://www.article19.org/wp-content/uploads/2021/01/ER-Tech-China-Report.pdf}
  {Emotional entanglement: China’s emotion recognition market and its
  implications for human rights}.

\bibitem[{Bender and Friedman(2018)}]{bender-friedman-2018-data}
Emily~M. Bender and Batya Friedman. 2018.
\newblock \href {https://doi.org/10.1162/tacl_a_00041} {Data statements for
  natural language processing: Toward mitigating system bias and enabling
  better science}.
\newblock \emph{Transactions of the Association for Computational Linguistics},
  6:587--604.

\bibitem[{Birhane(2021)}]{Birhane2021impossibility}
Abeba Birhane. 2021.
\newblock The impossibility of automating ambiguity.
\newblock \emph{Artificial Life}, 27(1):44--61.

\bibitem[{Bissoto et~al.(2020)Bissoto, Valle, and Avila}]{bissoto2020debiasing}
Alceu Bissoto, Eduardo Valle, and Sandra Avila. 2020.
\newblock Debiasing skin lesion datasets and models? not so fast.
\newblock In \emph{Proceedings of the IEEE/CVF Conference on Computer Vision
  and Pattern Recognition Workshops}, pages 740--741.

\bibitem[{Buolamwini and Gebru(2018)}]{buolamwini2018gender}
Joy Buolamwini and Timnit Gebru. 2018.
\newblock Gender shades: Intersectional accuracy disparities in commercial
  gender classification.
\newblock In \emph{Conference on fairness, accountability and transparency},
  pages 77--91. PMLR.

\bibitem[{Checco et~al.(2017)Checco, Roitero, Maddalena, Mizzaro, and
  Demartini}]{checco2017let}
Alessandro Checco, Kevin Roitero, Eddy Maddalena, Stefano Mizzaro, and Gianluca
  Demartini. 2017.
\newblock Let's agree to disagree: Fixing agreement measures for crowdsourcing.
\newblock In \emph{Proceedings of the Fifth AAAI Conference on Human
  Computation and Crowdsourcing}.

\bibitem[{Commission(2020)}]{european2020artificial}
European Commission. 2020.
\newblock On artificial intelligence—a european approach to excellence and
  trust.

\bibitem[{Fletcher-Watson et~al.(2018)Fletcher-Watson, De~Jaegher, Van~Dijk,
  Frauenberger, Magn{\'e}e, and Ye}]{fletcher2018diversity}
Sue Fletcher-Watson, Hanne De~Jaegher, Jelle Van~Dijk, Christopher
  Frauenberger, Maurice Magn{\'e}e, and Juan Ye. 2018.
\newblock Diversity computing.
\newblock \emph{Interactions}, 25(5):28--33.

\bibitem[{Floridi(2014)}]{floridi2014open}
Luciano Floridi. 2014.
\newblock Open data, data protection, and group privacy.
\newblock \emph{Philosophy \& Technology}, 27(1):1--3.

\bibitem[{Gebru et~al.(2018)Gebru, Morgenstern, Vecchione, Vaughan, Wallach,
  Daum{\'e}, and Crawford}]{Gebru2018DatasheetsFD}
Timnit Gebru, Jamie~H. Morgenstern, Briana Vecchione, Jennifer~Wortman Vaughan,
  H.~Wallach, Hal Daum{\'e}, and Kate Crawford. 2018.
\newblock Datasheets for datasets.
\newblock In \emph{Proceedings of the conference on Fairness, Accountability,
  and Transparency in Machine Learning}, Stockholm, Sweden.

\bibitem[{Kalluri(2020)}]{kalluri2020don}
Pratyusha Kalluri. 2020.
\newblock Don't ask if {A}rtificial {I}ntelligence is good or fair, ask how it
  shifts power.
\newblock \emph{Nature}, 583(7815):169--169.

\bibitem[{Kiritchenko and Mohammad(2018)}]{kiritchenko-mohammad-2018-examining}
Svetlana Kiritchenko and Saif Mohammad. 2018.
\newblock \href {https://doi.org/10.18653/v1/S18-2005} {Examining gender and
  race bias in two hundred sentiment analysis systems}.
\newblock In \emph{Proceedings of the Seventh Joint Conference on Lexical and
  Computational Semantics}, pages 43--53, New Orleans, Louisiana. Association
  for Computational Linguistics.

\bibitem[{McQuillan(2018)}]{mcquillan2018data}
Dan McQuillan. 2018.
\newblock Data science as machinic neoplatonism.
\newblock \emph{Philosophy \& Technology}, 31(2):253--272.

\bibitem[{Mitchell et~al.(2019)Mitchell, Wu, Zaldivar, Barnes, Vasserman,
  Hutchinson, Spitzer, Raji, and Gebru}]{mitchell2019model}
Margaret Mitchell, Simone Wu, Andrew Zaldivar, Parker Barnes, Lucy Vasserman,
  Ben Hutchinson, Elena Spitzer, Inioluwa~Deborah Raji, and Timnit Gebru. 2019.
\newblock Model cards for model reporting.
\newblock In \emph{Proceedings of the conference on fairness, accountability,
  and transparency}, pages 220--229.

\bibitem[{Mohammad(2022)}]{Mohammad21-Ethics-AER}
Saif~M. Mohammad. 2022.
\newblock Ethics sheet for automatic emotion recognition and sentiment
  analysis.
\newblock \emph{Computational Linguistics}.

\bibitem[{Ongweso(2020)}]{ongweso_2020}
Edward Ongweso. 2020.
\newblock \href
  {https://www.vice.com/en/article/g5pawq/an-ai-paper-published-in-a-major-journal-dabbles-in-phrenology}
  {An ai paper published in a major journal dabbles in phrenology}.

\bibitem[{Picard(2000)}]{picard2000affective}
Rosalind~W Picard. 2000.
\newblock \emph{Affective computing}.
\newblock MIT press.

\bibitem[{Prates et~al.(2019)Prates, Avelar, and Lamb}]{prates2019assessing}
Marcelo~OR Prates, Pedro~H Avelar, and Luis~C Lamb. 2019.
\newblock Assessing gender bias in machine translation: a case study with
  google translate.
\newblock \emph{Neural Computing and Applications}, pages 1--19.

\bibitem[{Schwartz et~al.(2020)Schwartz, Dodge, Smith, and
  Etzioni}]{schwartz2020green}
Roy Schwartz, Jesse Dodge, Noah~A Smith, and Oren Etzioni. 2020.
\newblock Green {AI}.
\newblock \emph{Communications of the ACM}, 63(12):54--63.

\bibitem[{Strubell et~al.(2020)Strubell, Ganesh, and
  McCallum}]{Strubell_Ganesh_McCallum_2020}
Emma Strubell, Ananya Ganesh, and Andrew McCallum. 2020.
\newblock \href {https://doi.org/10.1609/aaai.v34i09.7123} {Energy and policy
  considerations for modern deep learning research}.
\newblock \emph{Proceedings of the AAAI Conference on Artificial Intelligence},
  34(09):13693--13696.

\bibitem[{Talat et~al.(2021)Talat, Blix, Valvoda, Ganesh, Cotterell, and
  Williams}]{talat2021word}
Zeerak Talat, Hagen Blix, Josef Valvoda, Maya~Indira Ganesh, Ryan Cotterell,
  and Adina Williams. 2021.
\newblock \href {http://arxiv.org/abs/2111.04158} {A word on machine ethics: A
  response to jiang et al. (2021)}.

\bibitem[{Wakefield(2021)}]{wakefield_2021}
Jane Wakefield. 2021.
\newblock \href {https://www.bbc.com/news/technology-57101248} {Ai
  emotion-detection software tested on uyghurs}.

\bibitem[{Woensel and Nevil(2019)}]{woensel_nevil_2019}
Lieve~Van Woensel and Nissy Nevil. 2019.
\newblock \href
  {https://www.europarl.europa.eu/RegData/etudes/ATAG/2019/634415/EPRS_ATA(2019)634415_EN.pdf}
  {What if your emotions were tracked to spy on you?}

\end{thebibliography}
\bibliographystyle{acl_natbib}

\appendix

\section{FAQ and Discussion}
\label{sec:appendix}

\noindent {\it Q1. Should we create ethics sheets for a handful of AI Tasks (more prone to being misused, say) 
or do we need ethics sheets for all AI tasks?}\\[5pt] 
\noindent { A.} To me, the answer is clear. We need to create ethics sheets for every task that has significant impact on people
or deals with people or their artefacts in any significant way.
This follows from the idea that we need to think about ethics considerations pro-actively and not as a reaction to 
harms that we observe after system deployment. Different AI tasks may be more or less prone to controversy, but all AI tasks impact people in some way, and thus have ethical considerations. Sometimes even small and seemingly innocuous choices can have far-reaching implications. Sometimes a thoughtful consideration can help make a small, but notable difference, to improve someone's life.

Ethics sheets for AI Tasks can provide the means for us as a collective to provide, in writing,
what we think are the ethical considerations and the societal implications of AI Tasks.
For some tasks, this document can be short and straightforward indicating minimum risk;
and that document and the process that led to it are still useful. We do not know if there is minimum risk without
some amount of investigation. Also, 
\begin{quote}
    \textit{A written document allows others to challenge our assumptions and conclusions.} 
\end{quote}
This is a good thing!
We cannot predict everything and anticipate every harm. We should not let that stop us from creating a working document that will be useful to others. Ethics sheets will always be incomplete and require revisions. Periodically revising the document builds on our knowledge.\\[8pt]
\noindent {\it Q2. Who should create ethics sheets?}\\[5pt]
\noindent { A.} There are two things going on here:\\[-16pt]
\begin{enumerate}[itemsep=-1pt]
\item Who should take a \textit{lead} in developing ethics sheets (who takes on more of the burden)? 
\item Whose voices should be included? 
\end{enumerate}
\vspace*{-1mm}
\noindent For 1, anyone or any group can take the lead. Researchers already working on the task (or proposing a new task) are well-positioned to take the lead as they are familiar with the intricacies of the task and likely thinking about the ethical implications already. However, experienced researchers may have more blind spots. New researchers, especially those from Social Science, Psychology, Linguistics, etc.\@ can bring vital new insights.\\[2pt]
\noindent For 2, the goal is to include voices of all stakeholders (especially of those impacted by the technology).
However, the process can be iterative, starting at a smaller scale.\\[8pt]
\noindent {\it Q3. When should we create Ethics Sheets for AI Tasks?} Normally, we learn about ethical issues because/after they have been deployed.\\[5pt]
\noindent { A.} While we cannot foresee all consequences of our creations, it would be fair to say AI researchers have not done enough to anticipate the negative consequences of systems that we have created and deployed. Additionally, with great work over the last few years highlighting the ethical implications of AI systems, we are better placed to anticipate issues for the future. Therefore:\\[4pt]
\noindent \textit{For existing tasks:} create ethics sheets now; revisit and update periodically.\\[4pt]
\noindent {\it For new tasks:} create ethics sheets along with the paper introducing the task; as the task has more buy-in from the  community, others can also create a new ethics sheet or update the existing one. \\[8pt]
\noindent {\it Q4. Does it matter what we define as a `task'?} AI tasks can be defined at a high/general level (e.g., automatic emotion recognition) or fine/specific level (e.g., detecting sentiment in book reviews).\\[5pt]
\noindent { A.} We can let community interest and expertise guide what task definitions are used (similar to topics of survey papers). It is great to have multiple overlapping ethics sheets that cover AI tasks at overlapping levels of specificity. There is no ``objective'' or ``correct'' ethics sheet or survey article. There is no one ``correct'' scope or task definition for ethics sheets. It is useful to have multiple ethics sheets for the same or overlapping tasks, just as it is useful to have multiple survey articles for overlapping areas of research.\\[8pt] 
\noindent {\it Q5. Should the sheets depend on the kind of data or modality involved?}\\[5pt]
\noindent { A.} Yes, one can create focused ethics sheets as appropriate. In the example AER sheet, I  specify in the ``Scope and Modalities'' section that the sheet focuses primarily on AER from language (text).\\ [8pt]
\noindent {\it Q6. Should we think about research systems differently from deployed systems?}\\[5pt]
\noindent { A.} In my view, deployed systems have a much higher bar in terms of balancing the many ethical considerations. It is common for research systems to focus on a smaller number of dimensions (say accuracy on certain test sets) ignoring certain other dimensions. However, research systems are often picked up by developers and deployed. So research systems should make their dimensions of focus clear to the reader/user. They should also discuss the suitability of deploying such a system, intended uses, and ethical issues that may arise if one deploys their system.\\[8pt]
\noindent \textit{Q7. Why should academic researchers care about the ethics of system deployment?}\\[5pt]
\noindent A. Academic research feeds commercial research and development. We need to communicate the ethical considerations of what we create.
Also, we are often not in positions of conflict of interest;
no danger of losing our job for raising concerns.\\[8pt]
 \noindent \textit{Q8. Should ethics sheets be updated?} \\[5pt]
 \noindent A. Yes,
 as technologies change and as society embraces new values, 
 we need to create revisions or new sheets. Ethics sheets will act as an explicit record of what was considered important by different groups of people at different times.\\ [8pt]
 \noindent \textit{Q9. Won't ethics sheets slow things down?} \\[5pt]
 \noindent A. Ethics sheets aid in a win--win scenario: Assuming that one wants to create AI systems responsibly, having access to 
 one or more ethics sheets for their task will help a researcher/developer obtain their goal faster.
 Also, we do not want to be going fast at the expense of others. 
 Developing systems responsibly is in the best interest of all concerned.
 In that sense,
 slowing down is good. See this wonderful talk by Min-Yen Kan.\footnote{https://www.youtube.com/watch?v=hEK18EsDGzc}

\end{document}